\documentclass{ieeeaccess}
\usepackage{cite}
\usepackage{amsmath,amssymb,amsfonts}
\usepackage{algorithmic}
\usepackage{graphicx}
\usepackage{textcomp}

\usepackage{bm}
\makeatletter
\AtBeginDocument{\DeclareMathVersion{bold}
\SetSymbolFont{operators}{bold}{T1}{times}{b}{n}
\SetSymbolFont{NewLetters}{bold}{T1}{times}{b}{it}
\SetMathAlphabet{\mathrm}{bold}{T1}{times}{b}{n}
\SetMathAlphabet{\mathit}{bold}{T1}{times}{b}{it}
\SetMathAlphabet{\mathbf}{bold}{T1}{times}{b}{n}
\SetMathAlphabet{\mathtt}{bold}{OT1}{pcr}{b}{n}
\SetSymbolFont{symbols}{bold}{OMS}{cmsy}{b}{n}
\renewcommand\boldmath{\@nomath\boldmath\mathversion{bold}}}
\makeatother

\def\BibTeX{{\rm B\kern-.05em{\sc i\kern-.025em b}\kern-.08em
    T\kern-.1667em\lower.7ex\hbox{E}\kern-.125emX}}

\begin{document}
\history{Date of publication xxxx 00, 0000, date of current version xxxx 00, 0000.}
\doi{10.1109/ACCESS.2024.0429000}

\title{HiFiSeg: High-Frequency Information Enhanced Polyp Segmentation with Global-Local Vision Transformer}
\author{\uppercase{JINGJING REN}, 
\uppercase{XIAOYONG ZHANG}, and LINA ZHANG.
}

\address{Department of Intelligence and Information Engineering, Taiyuan University, Taiyuan 030032, CHINA (e-mail: renjingjing@tyu.edu.cn)}

\tfootnote{This work was supported in part by the Science and Technology Innovation Program of Higher Education Institutions in Shanxi Province 
2024L386}

\markboth
{JINGJING REN \headeretal: HiFiSeg: High-Frequency Information Enhanced Polyp Segmentation with Global-Local Vision Transformer}
{JINGJINGREN \headeretal: HiFiSeg: High-Frequency Information Enhanced Polyp Segmentation with Global-Local Vision Transformer}

\corresp{Corresponding author: JINGJING REN (e-mail: 2014010020@tyu.edu.cn).}

\begin{abstract}
Numerous studies have demonstrated the strong performance of vision transformer (ViT)-based methods across various computer vision tasks. However, ViT models often struggle to effectively capture high-frequency components in images, which are crucial for detecting small targets and preserving edge details, especially in complex scenarios. This limitation is particularly challenging in colon polyp segmentation, where polyps exhibit significant variability in structure, texture, and shape. High-frequency information, such as boundary details, is essential for achieving precise semantic segmentation in this context.
To address these challenges, we propose HiFiSeg, a novel network for colon polyp segmentation that enhances high-frequency information processing through a global-local vision transformer framework. HiFiSeg leverages the pyramid vision transformer (PVT) as its encoder and introduces two key modules: the global-local interaction module (GLIM) and the selective aggregation module (SAM). GLIM employs a parallel structure to fuse global and local information at multiple scales, effectively capturing fine-grained features. SAM selectively integrates boundary details from low-level features with semantic information from high-level features, significantly improving the model's ability to accurately detect and segment polyps.
Extensive experiments on five widely recognized benchmark datasets demonstrate the effectiveness of HiFiSeg for polyp segmentation. Notably, the mDice scores on the challenging CVC-ColonDB and ETIS datasets reached 0.826 and 0.822, respectively, underscoring the superior performance of HiFiSeg in handling the specific complexities of this task.
\end{abstract}


\titlepgskip=-21pt

\maketitle

\section{Introduction}
\label{sec:introduction}
\PARstart{P}{olyps} are abnormal growths in the colon and rectum that protrude from the intestinal mucosa. Colorectal cancer frequently arises from colonic polyps, especially adenomatous ones, making early detection and removal critical for preventing cancer progression. Colonoscopy is widely regarded as the gold standard for detecting colorectal lesions~\cite{favoriti2016worldwide}. However, the manual annotation of polyps during colonoscopy is both time-consuming and prone to human error, underscoring the need for automated and accurate image segmentation methods to assist in diagnosis.

Deep learning algorithms, particularly CNNs, have achieved significant success in medical image applications such as cardiac, skin lesion, and polyp segmentation~\cite{yan2024enhancing}. Fully convolutional networks (FCNs), including models like UNet~\cite{ronneberger2015u}, SegNet~\cite{badrinarayanan2017segnet}, and DeepLab~\cite{chen2017deeplab}, have become the dominant approaches in this domain. However, due to the limited receptive fields of CNNs, these methods struggle to capture long-range dependencies and global context, which are essential for accurately representing shape and structural information in medical image segmentation.

\begin{figure}[t] % t 选项将图片浮动到页面顶部
  \centering
  \includegraphics[width=1\linewidth]{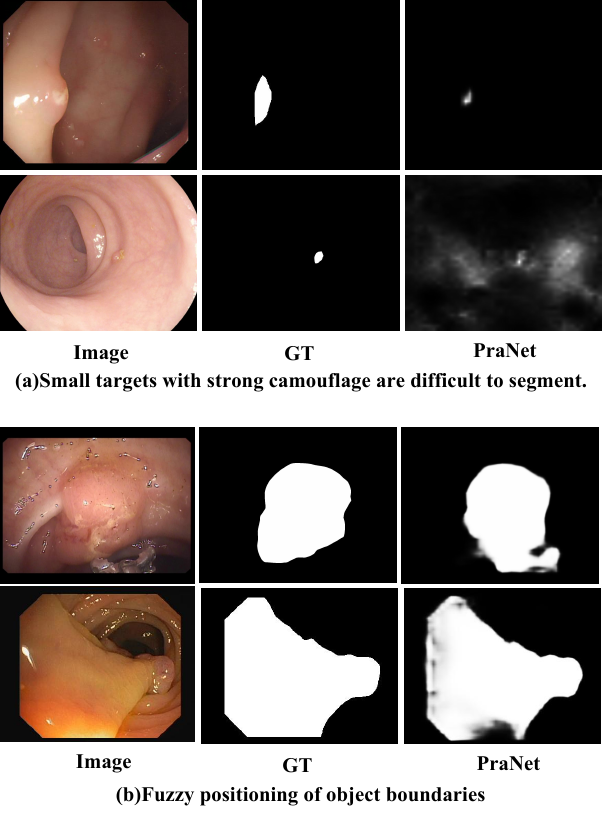}
  \vspace{-15pt}
  \caption{Examples of segmentation of the PraNet model for different challenge cases.}
  \label{fig:1}
\end{figure}

The transformer~\cite{vaswani2017attention} architecture, with its multi-head self-attention (MHSA) mechanism, excels at capturing complex spatial transformations and long-range dependencies. While it has seen tremendous success in natural language processing (NLP), its adaptation to vision tasks through the vision transformer (ViT)~\cite{dosovitskiy2020image} was aimed at overcoming the limitations of CNNs in image recognition. However, despite Transformers' ability to model global dependencies, they struggle with capturing image locality and maintaining translational invariance, which is critical for accurately segmenting small targets and boundaries.
To address this challenge, recent works have proposed hybrid architectures~\cite{shen2023git, shen2023pedestrian, shen2023triplet} that combine the strengths of both Transformers and CNNs, such as TransUnet~\cite{chen2021transunet}, HiFormer~\cite{heidari2023hiformer}, and LeVit-Unet~\cite{xu2023levit}. These models aim to leverage the locality of CNNs with the global context captured by Transformers, enabling them to encode both local and global features for medical image segmentation. While these hybrid models have shown improved performance, they still face limitations, particularly in capturing fine-grained details. This shortcoming affects the accurate identification of small targets and boundary localization, hindering the model's ability to generalize effectively in medical image segmentation.
As illustrated in Figure \ref{fig:1}, the PraNet model highlights some of the persistent challenges in polyp segmentation, underscoring the need for better methods to address these issues.

\begin{figure*}[t] % t 选项将图片浮动到页面顶部
  \centering
  \includegraphics[width=0.8\textwidth]{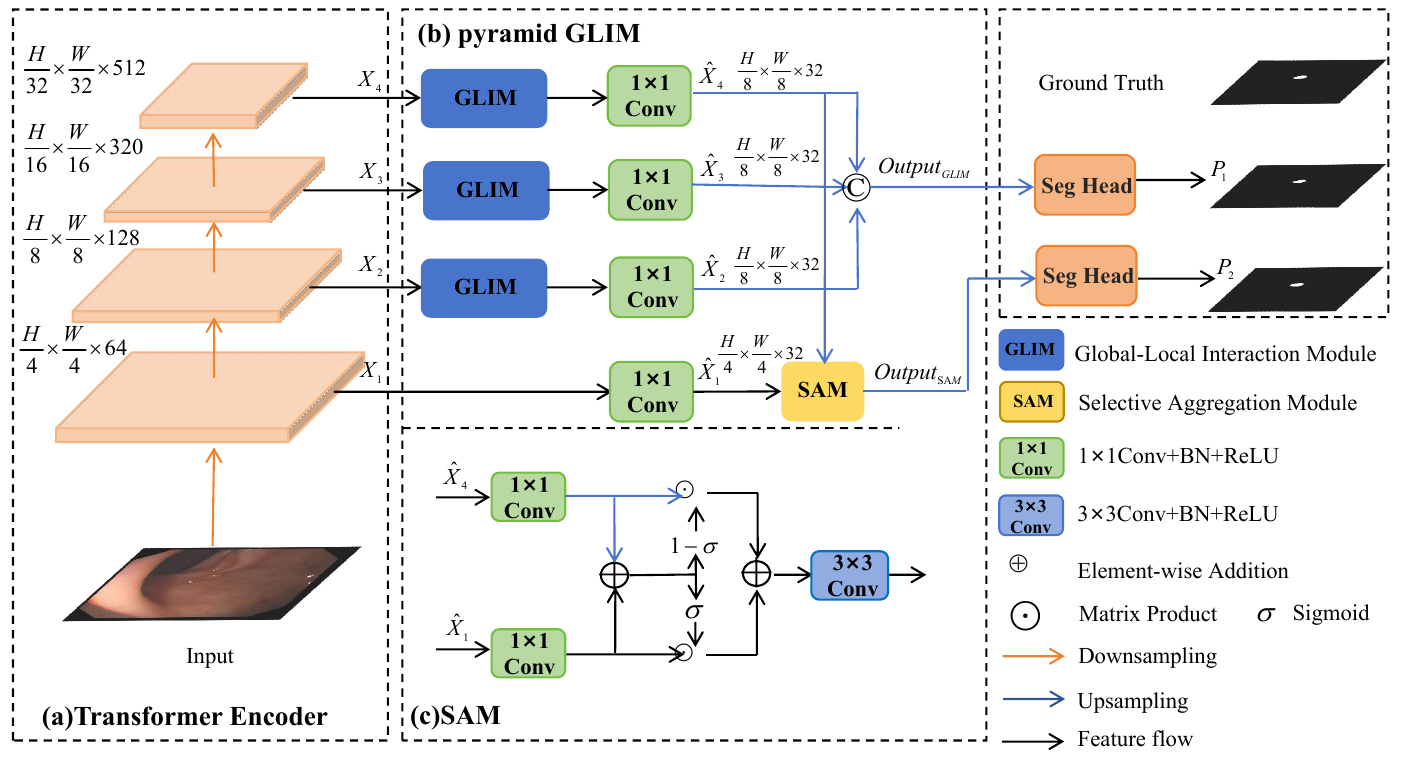}
  %\vspace{pt}
  \caption{The overall architecture of HiFSeg network. The entire model contains three components: (a) pyramid vision transformer (PVT) as encoder; (b) pyramid global-local interaction module(GLIM)
for fusing multi-level features; (c) selective aggregation module(SAM) for integrating the high- and low-level features selectively for the final output.}
  \label{fig:2}
\end{figure*}

Inspired by multiscale and multilevel feature modeling approaches~\cite{wang2021pyramid,gao2024res2net,zhang2023robust,wang2023mca}, we propose a high-frequency information-enhanced polyp segmentation framework, termed \textbf{HiFiSeg}. The main components of HiFiSeg include the pyramid vision transformer (PVT), the global-local interaction module (GLIM), and the selective aggregation module (SAM).  PVT, a lightweight hierarchical Transformer, serves as the encoder to capture multiscale features efficiently. GLIM employs parallel convolutional kernels and pooling operations of varying sizes to aggregate global and local information, allowing the extraction of fine-grained features. This is particularly advantageous for localizing small targets. To reduce computational complexity, GLIM uses grouped channels with depthwise separable convolution. SAM refines boundary features by leveraging high-level semantic information to guide the selective refinement of low-level details. 

In summary, our contributions are as follows:

\textbullet\ We propose \textbf{HiFiSeg}, a novel framework for colon polyp segmentation. HiFiSeg utilizes the Pyramid Vision Transformer as an encoder to capture more robust features than CNN-based methods.

\textbullet\ We design two key modules, \textbf{GLIM} and \textbf{SAM}, to enhance the framework. GLIM improves segmentation performance for small targets by extracting multiscale local features, while SAM addresses boundary ambiguity by selectively fusing low-level boundary details with high-level semantic information.

\textbullet\ We evaluate HiFiSeg on five standard benchmark datasets for polyp segmentation, including Kvasir~\cite{jha2020kvasir}, CVC-ClinicDB~\cite{bernal2015wm}, CVC-300~\cite{vazquez2017benchmark}, CVC-ColonDB~\cite{tajbakhsh2015automated}, and ETIS~\cite{silva2014toward}. On the challenging CVC-ColonDB and ETIS datasets, HiFiSeg achieves mDice scores of 0.826 and 0.822, respectively, surpassing existing state-of-the-art methods.

\section{Related Work}
\subsection{Convolutional Neural Networks}  %和任务相关

CNNs are deep learning models specifically designed for processing image data, excelling in feature extraction capabilities, and are widely used in computer vision tasks. In recent years, CNN-based structures represented by the UNet architecture have made significant progress in medical image segmentation. UNet consists of a symmetric encoder and decoder, with skip connections that transfer features from the encoder to the decoder, combining low-level features and high-level semantic information to achieve high-precision segmentation. Many works have made improvements based on the UNet architecture, such as UNet++~\cite{zhou2018unet++}, ResUNet++~\cite{jha2019resunet++} and DoubleUnet ~\cite{jha2020doubleu}. 

Unlike the UNet-based methods, PolypNet~\cite{banik2020polyp} proposed a dual-tree wavelet pooling CNN with a local gradient-weighted embedding level set, significantly reducing the false positive rate, significantly reducing the false positive rate. Caranet~\cite{lou2022caranet} proposed a context axial reserve attention network to improve the segmentation performance on small objects. PraNet~\cite{fan2020pranet} generates a global map based on high-level features aggregated by the parallel partial decoder and employs the reverse attention module to mine boundary cues, effectively correcting any misaligned predictions, thereby improving segmentation accuracy.
\begin{figure}[t]
    \centering
    \includegraphics[width=1\linewidth]{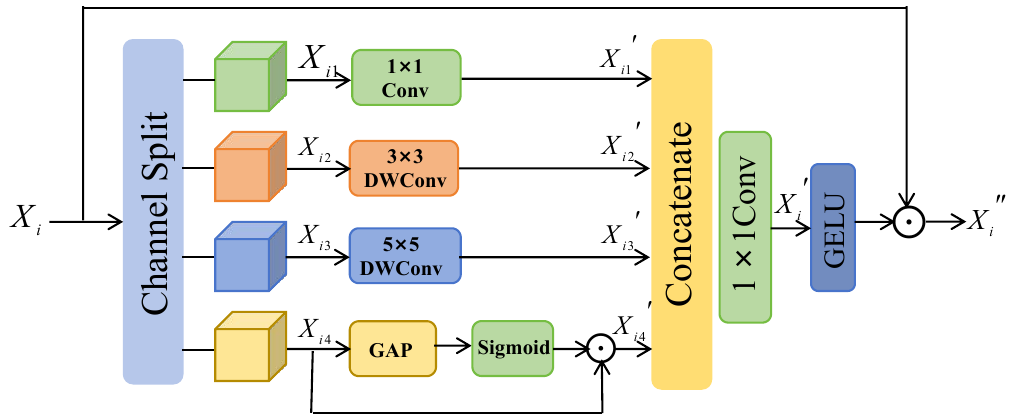}
\vspace{-15pt}
\caption{Details of the introduced global-local interaction module(GLIM).It consists of three convolutional branches and a global average pooling branch.}
\vspace{0.1cm}
\label{fig:3}
\end{figure}
\subsection{Vision Transformer}  % 和你的AB模块技术相关
Transformer~\cite{vaswani2017attention}, proposed by Vaswani et al., uses multi-head self-attention to capture long-range dependencies. Initially designed for natural language tasks like translation, Transformers are now widely used in image processing and speech recognition due to their parallel processing and global context modeling. Vision transformer (ViT)~\cite{dosovitskiy2020image} was the first pure Transformer for image classification, processing images as fixed-size patches. Subsequent models like Swin Transformer~\cite{liu2021swin}, PVT~\cite{wang2021pyramid}, and Segformer~\cite{xie2021segformer} introduced pyramid structures for improved vision tasks.
Meanwhile, diffusion models~\cite{shen2023advancing, shen2024boosting, shen2024imagdressing, imagpoes} have become popular for iteratively refining images through noise reduction. When combined with Transformers, they enhance feature extraction and segmentation, improving performance in tasks like medical image segmentation and object detection.

In medical image segmentation, hybrid architectures combining Transformers and CNNs have shown promise. Transfuse~\cite{zhang2021transfuse} integrates Transformers and CNNs to capture global and local features, while TransUNet~\cite{chen2021transunet} uses Transformers as encoders and U-Net to refine local details. Polyp-PVT~\cite{dong2021polyp} leverages PVT as an encoder with a graph-based similarity aggregation module. ColonFormer~\cite{duc2022colonformer} models global semantic relations and refines polyp boundaries, while DuAT~\cite{tang2023duat} introduces dual-aggregate Transformers to balance large and small target detection and enhance boundary precision.

\section{Proposed Method}
\subsection{Overview}
As shown in Figure \ref{fig:2}, our proposed network HiFiSeg consists of a pyramid vision transformer (PVT) encoder, global-local interaction module(GLIM), and selective aggregation module(SAM). The PVT encoder is employed to extract multi-scale hierarchical features from the input image \newcommand{\HW}{H \times W}
$X \in \mathbb{R}^{\HW \times 3} $, capturing both fine-grained local details and broad semantic information. Specifically, the PVT backbone yields four pyramid features $X_{i} \in \mathbb{R}^{\frac{H}{2^{i+1}} \times \frac{W}{2^{i+1}} \times C_{i}}, \text{ where } i \in \{1,2,3,4\} \text{ and } C_{i} \in \{64,128,320,512\}$. The high-level features ${\{X_{i}|i \in\ (2, 3, 4)\}}$ are fed into the GLIM module to extract local multiscale features. The outputs of the GLIM module are then concatenated to produce the fused global-local multiscale feature $Output_{GLIM}$. The low-level feature $X_{1}$ is selectively aggregated with the high-level feature $X_{4}$  through the SAM module to obtain the enhanced edge feature $Output_{SAM}$.
Finally, $Output_{GLIM}$ and $Output_{SAM}$ are fed into the segmentation heads to obtain the predicted  results $P_{1}$ and $P_{2}$, respectively.

\subsection{Transformer Encoder}  %A模块名称介绍，一定是用公式字母介绍。假设X，a b c 分别是什么
Some recent studies have demonstrated that pyramid structures, through the integration of multi-scale contextual information, can substantially improve the accuracy and efficiency of image segmentation. Our model uses the pyramid vision transformer (PVT) proposed in [46] as the encoder backbone to extract more robust features for polyp segmentation. PVT is the first pure Transformer backbone designed for various pixel-level dense prediction tasks. In polyp segmentation, PVT generates four multi-scale feature maps ${\{X_{i}|i \in\ (1, 2, 3, 4)\}}$. Among these feature maps, $X_{1}$ gives detailed information about the polyps, while $X_{2}, X_{3}$, and $X_{4}$ provide high-level features. 
\begin{figure*}[ht] % t 选项将图片浮动到页面顶部
  \centering
  \includegraphics[width=\textwidth]{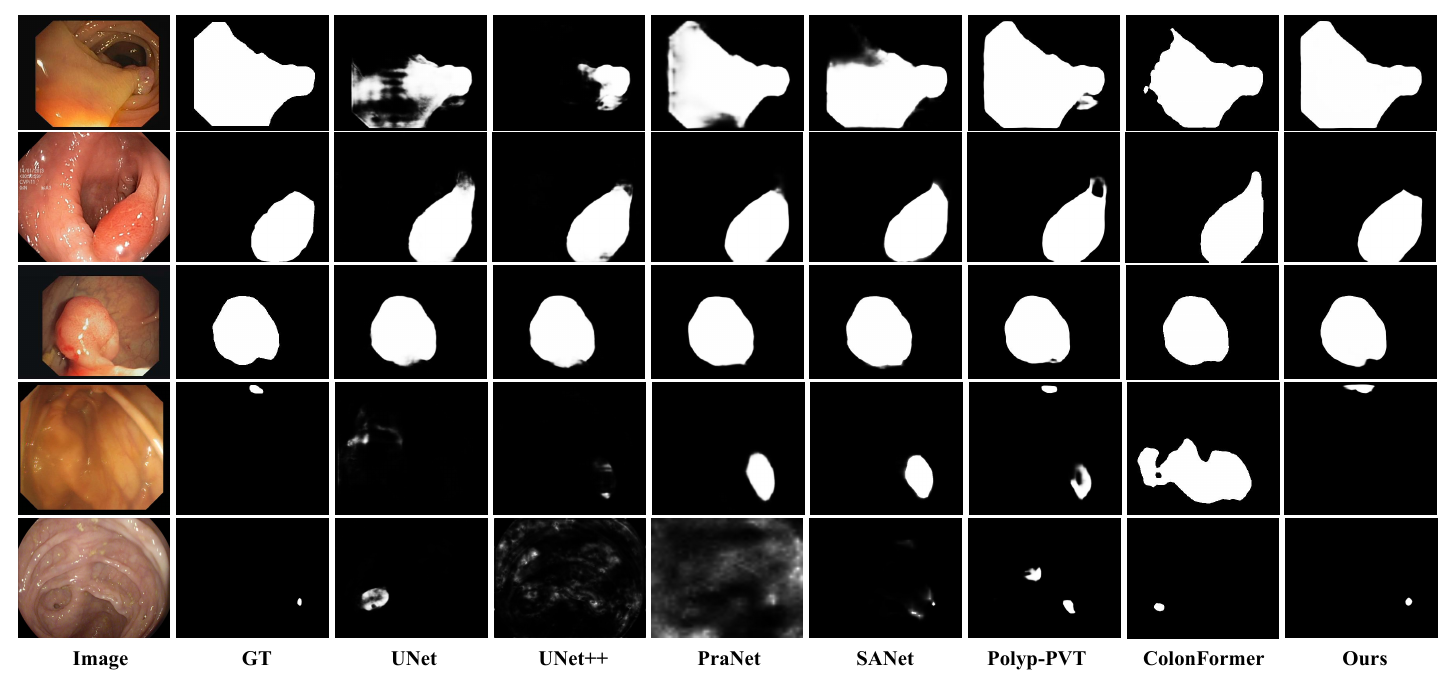}
  \vspace{-15pt} % 调整标题与图像的距离
  \caption{Qualitative results comparison of different models.}
  \label{fig:4}
\end{figure*}
\subsection{Global-Local Interaction Module}  %B模块介绍，定是用公式字母介绍。假设X，a b c 分别是什么
In medical image segmentation, the context and background often occupy a much larger area than the segmentation target itself. Consequently, capturing information across different scales is essential for accurately segmenting small targets. Instead of presenting multi-scale features in a layer-by-layer fashion, GLIM aggregates global and local features at a specific high-level, achieving multi-scale feature representation at a granular level, reducing errors in high-level features.

The detailed architecture of our propoed GLIM is depicted in Figure \ref{fig:3}, consisting of three convolution branches and one pooling branch. To balance accuracy and computational resources, we evenly divide the channels into four groups, applying depthwise separable convolution for each. After splitting the input features ${\{X_{i} \mid i \in \{2, 3, 4\}\}}$
 into four components ${X_{i1},X_{i2},X_{i3},X_{i4}}$, they are fed into feature generation units at different scales. Given the input feature $X_{i}$, this procedure can be formulated as:
\begin{equation}
\begin{aligned}
    X_{i1}&, X_{i2}, X_{i3}, X_{i4} = Split(X)\\
    X^{'}_{i1} &= Conv_{1\times1}(X_{i1})\\
    X^{'}_{i2} &= DWConv_{3\times3}(X_{i2})\\
    X^{'}_{i3} &= DWConv_{5\times5}(X_{i3})\\
    X^{'}_{i4} &= \text{Sigmoid}(\text{GAP}(X_{i4})) \odot (X_{i4}),
\end{aligned}
\tag{1}
\end{equation}
where $Split(\cdot)$ denotes the channel separation operation, $Conv_{1\times1}(\cdot)$ represents 1×1 convolution, $DWConv_{3\times3}(\cdot)$ refers to 3×3 depth-wise convolution, $DWConv_{5\times5}(\cdot)$ refers to 5×5 depth-wise convolution, GAP stands for global average pooling, Sigmoid refers to the Sigmoid activation function, and $\odot$ is the
element-wise product. The convolution branchs employ kernels of varying sizes to extract features at different scales of the image, while the pooling branch uses global average pooling to aggregate global information. These feature maps ${\{X^{'}_{i1},X^{'}_{i2},X^{'}_{i3},X^{'}_{i4}}\}$ are then concatenated along the channel dimension, and a 1×1 convolution is applied to aggregate both global and local information, resulting in a rich feature representation $X^{'}_{i}$.This process can be expressed as:
\begin{equation}
X^{'}_{i} = Conv_{1\times1}(Concat([X^{'}_{i1},X^{'}_{i2},X^{'}_{i3},X^{'}_{i4}])),\tag{2}
\end{equation}
where $Concat(\cdot)$ denotes a concatenation operation, while $Conv_{1\times1}(\cdot)$ refers to a 1×1convolution. To enhance feature selection, we apply the GELU activation function to the feature $X^{'}_{i}$ to generate the attention feature map, and then modulate the input feature $X$ through element-wise multiplication. It can be formulated as:
\begin{equation}
X^{''}_{i} =GELU(X^{'}_{i})\odot X_{i}, \tag{3}
\end{equation}
where GELU refers to the GELU activation function, and $\odot$ is the
element-wise product.

\subsection{Selective Aggregation Module}  
Shallow features contain rich spatial information, while deep features contain more semantic information. The effective combination of these two is crucial for improving the accuracy of the model. In order to enhance the guidance of shallow detail features by deep semantic features, we propose the selective aggregation module (SAM), as shown in Figure 1(c). Unlike previous fusion methods that directly add the provided feature maps, SAM selectively aggregates the features. First, the shallow feature $\hat{X}_{1}$ and deep feature $\hat{X}_{4}$ are individually processed through 1×1 convolutions followed by sigmoid activations to produce the attention weight $\sigma$. The output of the Sigmoid function could be represented as:
\begin{equation}
\sigma=Sigmoid(C_{1\times 1}(\hat{X}_{1})\oplus C_{1\times 1}(\hat{X}_{4})). \tag{4}
\end{equation}

If $\sigma$ is high, the model assigns greater trust to the shallow feature, and vice versa.The output of the SAM can be written as:
\begin{equation}
Output_{SAM}=\sigma \hat{X}_{1}\oplus (1-\sigma)\hat{X}_{4}.
\tag{5}
\end{equation}

\subsection{Loss Function}  %看页数需要不够就加，页数够了就不要，可选。
We use weighted binary cross-entropy(BCE) loss and the weighted intersection over union(IoU) loss for supervision. Our loss function can be formulated as Eqn. 6:
\begin{equation}
\begin{aligned}
    L_{total}&=L({P_{1},G})+L({P_{2},G})\\
    L({P_{1},G})&=\lambda_{1} L^w_{BCE}(P_{1},G)+\lambda_{2} L^w_{IoU}(P_{1},G)\\
    L({P_{2},G})&=\lambda_{1} L^w_{BCE}(P_{2},G)+\lambda_{2} L^w_{IoU}(P_{2},G),
\end{aligned}
\tag{6}
\end{equation}
where $P_{1}$, $P_{2}$ are the outputs and G is the ground truth, $\lambda_{1}$ and $\lambda_{2}$ are the weighting coefficients, $L^w_{BCE}(\cdot)$ and $L^w_{IoU}(\cdot)$ are the  weighted BCE and weighted IoU.
\section{Experiment and Analysis}\label{sec:exp}  %xxx换成你的task, %yyy 换成数据集名称

To validate the proposed HiFiSeg method's superiority, it is compared with multiple state-of-the-art approaches on five popular datasets for polyp segmentation, namely, Kvasir~\cite{jha2020kvasir}, CVC-ClinicDB~\cite{bernal2015wm}, CVC-300~\cite{vazquez2017benchmark}, CVC-ColonDB~\cite{tajbakhsh2015automated}, ETIS~\cite{silva2014toward}.
\begin{table*}[t]
\centering

\caption {Quantitative Evaluation of the Learning Ability of Our Proposed HiFiSeg and Other Methods on the Kvasir and CVC-ClinicDB Datasets.}
\renewcommand{\arraystretch}{1.5}
\begin{tabular}{|p{0.15\textwidth}|l|c|c|c|c|c|}
% 定义每列的对齐方式：l(left), c(center), r(right)
\hline % 画水平线
{Methods} & \multicolumn{3}{c|}{Kvasir} & \multicolumn{3}{c|}{CVC-ClinicDB} \\ \cline{2-7}
& mDice & mIoU & MAE & mDice & mIoU & MAE \\ \hline
U-Net~\cite{ronneberger2015u}  & 0.818 & 0.746 & 0.055 & 0.823 & 0.755 &0.055\\ 
UNet++~\cite{zhou2018unet++} & 0.821 & 0.743 & 0.048 & 0.794 & 0.729 & 0.022 \\ 
Pranet~\cite{fan2020pranet} & 0.898 & 0.840 &0.030 & 0.899 & 0.849 &0.009 \\ 
SANet~\cite{wei2021shallow} & 0.904 & 0.847 & 0.028 & 0.916 & 0.859 & 0.012\\ 
TransUnet~\cite{chen2021transunet} &0.913 & 0.857 &0.028 &0.935 & 0.887 &0.008 \\
SSFormer~\cite{wang2022stepwise} & 0.926 & 0.874 & 0.023 & 0.927 & 0.876 & 0.009 \\
Polyp-PVT~\cite{dong2021polyp} & 0.917 & 0.864 & 0.023 & 0.937 & 0.889 & 0.006 \\
ColonFormer-S~\cite{duc2022colonformer} & 0.927 & 0.877&0.021 &0.932&0.883&0.008 \\
\hline
\textbf{HiFiSeg(Ours)}& \textbf{0.933} & \textbf{0.886}&\textbf{0.018} & \textbf{0.942}& \textbf{0.897}&\textbf{0.006}\\
\hline
\end{tabular}
\label{table1}
\end{table*}
\begin{table*}
\centering
\caption{Quantitative Evaluation of the Generalization Ability of Our Proposed HiFiSeg and Other Methods on the CVC-300, CVC-ColonDB and ETIS Datasets.}
\renewcommand{\arraystretch}{1.5}
\begin{tabular}{|p{0.15\textwidth}|l|c|c|c|c|c|c|c|c|}% 定义每列的对齐方式：l(left), c(center), r(right)

\hline % 画水平线
{Methods} & \multicolumn{3}{c|}{CVC-300} & \multicolumn{3}{c|}{CVC-ColonDB}& \multicolumn{3}{c|}{ETIS}\\ \cline{2-10}
 & mDice & mIoU & MAE & mDice & mIoU & MAE & mDice & mIoU & MAE \\ \hline
UNet~\cite{ronneberger2015u} & 0.710 & 0.627 & 0.022 & 0.512 & 0.444 & 0.061& 0.398 & 0.335 & 0.036\\
UNet++~\cite{zhou2018unet++}  & 0.707 & 0.624 & 0.018 &0.794 & 0.729 & 0.022 &0.401 & 0.344 & 0.035\\ 
Pranet~\cite{fan2020pranet}  & 0.851 & 0.797 &0.010 & 0.712 & 0.640 &0.043 & 0.628 & 0.567 &0.031 \\ 
SANet~\cite{wei2021shallow} & 0.888 & 0.815 & 0.008 &0.753 & 0.670 & 0.043 &0.750 & 0.654 & 0.015\\ 
TransUnet~\cite{chen2021transunet}  &0.893 & 0.660 &0.009 &0.781 & 0.699 &0.036 &0.731 & 0.824 &0.021\\
SSFormer~\cite{wang2022stepwise} & 0.887 & 0.821 & 0.007 & 0.772 & 0.697 & 0.036 & 0.767 & 0.698 & 0.016\\
Polyp-PVT~\cite{dong2021polyp} & 0.900 & 0.833 & 0.007  & 0.808 & 0.727 & 0.031 & 0.787 & 0.706 & 0.013\\
ColonFormer-S~\cite{duc2022colonformer} &0.894&0.826&0.008 &0.811&0.730&0.027 &0.789&0.711&0.013\\
\hline
\textbf{HiFiSeg (Ours)} &\textbf{0.905}&\textbf{0.839}&\textbf{0.006}&\textbf{0.826} & \textbf{0.749}&\textbf{0.028}&\textbf{0.822}&\textbf{0.743}&\textbf{0.012}\\
\hline
\end{tabular}
\label{table2}
\end{table*}
\subsection{Datasets}%数据集介绍要求每个数据集写的长度差不多
We used five challenging public datasets for the polyp segmentation task, including Kvasir~\cite{jha2020kvasir}, CVC-ClinicDB~\cite{bernal2015wm}, CVC-300~\cite{vazquez2017benchmark}, CVC-ColonDB~\cite{tajbakhsh2015automated}, and ETIS~\cite{silva2014toward}, to validate the learning and generalization capabilities of our model. Details for each dataset are as follows:

\textbf{Kvasir dataset:} The dataset consists of 1000 images with different resolutions from 720 × 576 to 1920 × 1072 pixels.

\textbf{CVC-ClinicDB dataset:}The dataset contains 612 polyp images which are extracted from 29 different endoscopic video clips.The resolution of  images is 384 x 288.

\textbf{CVC-300 dataset:}The dataset consists of 60 polyp images and the resolution of the images is 574 x 500.

\textbf{CVC-ColonDB dataset:}The dataset consists of 380 polyp images and the resolution of the images is 570 x 500.

\textbf{ETIS dataset:}The dataset consists of 196 polyp images and the resolution of the images is 1225 x 966.

\subsection{Evaluation Metrics}  %用的什么指标，页数不够写详细，页数够了缩略写

We employ three widely-used met-
rics in the field of medical image segmentation,i.e., mean Dice (mDice), mean IoU (mIoU) and mean
absolute error (MAE) to evaluate the model performances. Mean Dice and IoU are widely utilized metrics that primarily focus on assessing the internal consistency of segmentation results. MAE, on the other hand, measures the pixel-level accuracy by calculating the average absolute error between the predicted and actual values.

\subsection{Implementation Details} %写一些超参数，输入大小 优化器， 学习率， 数据增强策略，等等
We randomly split the images from Kvasir and CVC-ClinicDB into 80$\%$ for training and 20$\%$ for testing. And test on CVC-300, CVC-ColonDB and ETIS datasets. Due to the uneven resolution of the images, we resized them to 352×352 resolution.

We implement the HiFiSeg using the PyTorch framework, utilizing an NVIDIA RTX 3090 GPU. To enhance the model's robustness concerning varying image sizes, the training images are scaled by factors of 0.75, 1, and 1.25~\cite{duc2022colonformer}, respectively, before being fed into the model for learning. PVT encoder uses the same parameters as pvt\_v2\_b2~\cite{wang2022pvt}. The model is trained end-to-end using the AdamW~\cite{loshchilov2017decoupled} optimizer, with the learning rate and weight decay set to 1e-4. The batch size is configured to 16.
\renewcommand{\arraystretch}{1.5} % 调整表格的行间距
\begin{table*}[t]
\centering
\caption{QUANTITATIVE RESULTS FOR ABLATION STUDIES.}
\begin{tabular}
{|p{0.15\textwidth}|l|c|c|c|c|c|c|c|c|c|c|}% 定义每列的对齐方式：l(left), c(center), r(right)

\hline % 画水平线
{Methods} & \multicolumn{2}{|c|}{Kvasir}& \multicolumn{2}{c|}{CVC-ClinicDB}& \multicolumn{2}{c|}{CVC-300} & \multicolumn{2}{c|}{CVC-ColonDB}& \multicolumn{2}{c|}{ETIS}\\ \cline{2-11}
 & mDice & mIoU  & mDice & mIoU & mDice & mIoU  & mDice & mIoU & mDice & mIoU\\ \hline
Baseline & 0.910 & 0.856 & 0.903 & 0.847 & 0.869 & 0.792& 0.796 & 0.707 & 0.759 &0.668 \\
\hline
w/o GLIM  & 0.924 & 0.878 & 0.937 &0.892 & 0.892 & 0.828 &0.813 & 0.734 & 0.798 &0.725\\ 
w/o SAM  & 0.918 & 0.871 &0.924 & 0.879 & 0.896 &0.833 & 0.798 & 0.721 &0.801&0.729\\ 
w/o Conv  & 0.923 &0.877 &0.931&0.883 &0.886& 0.816&0.808&0.732 & 0.788&0.710\\ 
w/o GAP  & 0.928 & 0.878 & 0.934 & 0.884 & 0.899 &0.833 & 0.811 &0.726  &0.781  &0.704\\ 
\hline
\textbf{HiFiSeg} &\textbf{0.933}& \textbf{0.886}&\textbf{0.942} &\textbf{0.897}&\textbf {0.905}&\textbf{0.839}&\textbf{0.826}&\textbf{0.749}&\textbf{0.822}&\textbf{0.743}\\
\hline
\end{tabular}
\label{table3}
\end{table*}
\begin{figure*}[ht] % t 选项将图片浮动到页面顶部
  \centering
  \includegraphics[width=0.8\linewidth]{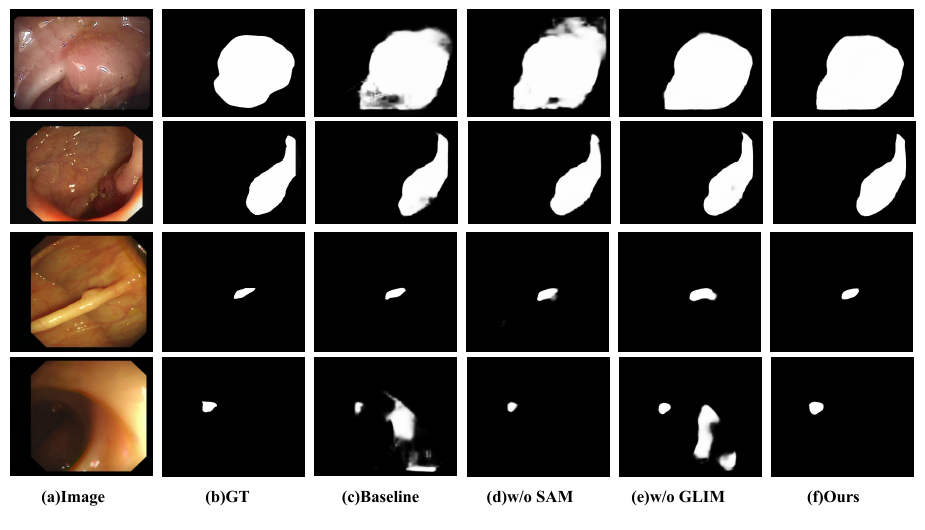}
  \vspace{-10pt}
  \caption{Visualization of the ablation study results.}
  \label{fig:5}
\end{figure*}
\subsection{Comparison with State-of-the-art Methods} %有几个数据集你就比较几个段落.记得课上讲的3句法。结论，举例，原因。
\subsubsection{Learning Ability}
We first evaluate the learning ability of the proposed model HiFiSeg on the training datasets Kvasir and ClinicDB. As shown in Table \ref{table1}, we compare our proposed HiFiSeg with recently published and classical models for polyp segmentation, including CNN-based models such as UNet~\cite{ronneberger2015u}, UNet++~\cite{zhou2018unet++}, PraNet~\cite{fan2020pranet}, and SANet~\cite{wei2021shallow}, as well as Transformer-based models like TransUnet~\cite{chen2021transunet}, SSFormer~\cite{wang2022stepwise}, Polyp-PVT~\cite{dong2021polyp}, and ColonFormer~\cite{duc2022colonformer}. These results demonstrate the effectiveness of our model in accurately segmenting polyps. Specifically, the HiFiSeg model has a mDice value of 0.933 and a mIoU value of 0.876 on Kvasir dataset, which are 0.6\% and 0.9\% higher than the best performing model, ColonFormer, respectively. For CVC-ClinicDB dataset, the HiFiSeg model has a mDice value of 0.942 and a mIoU value of 0.897, which are 0.6\% and 0.8\% higher than the best performing model, Polyp-PVT, respectively. 
\subsubsection{Generalization Capabilities}
To further evaluate our model's generalization performance, we test HiFiSeg on three unseen datasets: CVC-300, CVC-ColonDB, and ETIS. These datasets originate from different medical centers, each presenting unique challenges and characteristics. As seen in Table \ref{table2},
on three unseen datasets, our model outperforms peer models across all metrics, demonstrating strong generalization performance. On CVC-300 dataset, HiFiSeg achieves mDice of 0.905 and mIoU of 0.839, outperforming the second-best model, Polyp-PVT, by 0.5\% and 0.6\%, respectively. On CVC-ColonDB dataset, 
our model's mDice and mIoU scores are 1.5\% and 1.9\% higher than those of ColonFormer, respectively. Moreover, HiFiSeg achieves mDice of 0.822 and mIoU of 0.743 on ETIS dataset, which are 3.3\% and 3.2\% higher than the second-best model ColonFormer.
\subsubsection{Visual Results}
Figure \ref{fig:4} presents the visualization results of our model alongside the comparison models, providing a qualitative assessment of their performance. As shown in Figure \ref{fig:4}, our model produces significantly fewer incorrectly predicted pixels in the segmentation results compared to other models. It accurately identifies colonic tissues and polyps, efficiently captures the boundaries of tiny polyps and target objects, and maintains stable recognition and segmentation capabilities across various imaging conditions. As seen in the first three rows of Figure \ref{fig:4}, HiFiSeg accurately captures the boundaries and fine details of the target object, whereas the other methods fail to clearly detect the boundaries. In rows 4 and 5, our method demonstrates superior ability in identifying small targets and produces more accurate segmentation predictions.
\subsection{Ablation Studies and Analysis} 
We use PVTv2 as our baseline (Bas.) and evaluate module effectiveness by removing components from the complete GLIM. The training, testing, and hyperparameter settings are the same as mentioned in Sec. III-C. The results are shown in Table \ref{table3}.

\textbf{Effectiveness of GLIM.} To evaluate the effectiveness of GLIM, we trained a version of the model: "HiFiSeg (w/o GLIM)."As shown in Table \ref{table3}, compared to the standard HiFiSeg network, the performance of HiFiSeg (w/o GLIM) is reduced across all five datasets. This is particularly noticeable on ETIS dataset, where the mDice drops from 0.822 to 0.798 and the mIoU decreases from 0.743 to 0.725. As shown in the visualization results in Figure \ref{fig:5}, the HiFiSeg (w/o GLIM) model struggles to effectively distinguish between polyps and colon tissues and has difficulty accurately localizing targets, particularly small ones. In contrast, the HiFiSeg model, with the inclusion of the GLIM module, significantly improves the accuracy of target localization and small target detection due to the aggregation of local and global features.

\textbf{Effectiveness of SAM.} To evaluate the effectiveness of SAM, we trained a version of the model: "HiFiSeg (w/o SAM)."
As shown in Table \ref{table3}, compared to HiFiSeg(w/o SAM), HiFiSeg shows a substantial improvement in performance on all five datasets. Specifically, the mdice on CVC-ColonDB dataset is improved from 0.798 to 0.826 and the mIoU is mentioned from 0.721 to 0.749, both of which are improved by 2.8\%. The visualization results in Figure \ref{fig:5} show that SAM enables more accurate boundary extraction by effectively combining local pixel information with global semantic cues.

\textbf{Effectiveness of GLIM module components.} The GLIM consists of three convolution branches and one global average pooling(GAP) branch. The convolutional branches extract local features at multiple scales using convolutional kernels of varying sizes, while the global average pooling branch captures global information by spatially averaging the entire feature map, thereby better capturing the overall semantic context. To verify the effectiveness of the convolutional branches, we removed it from GLIM, resulting in HiFiSeg (w/o Conv). As shown in Table \ref{table3}, compared to the original HiFiSeg, the performance of the modified model drops significantly due to the lack of rich local representations, particularly on the ETIS dataset, where mDice and mIoU decrease by 3.4\% and 3.3\%, respectively. To verify the effectiveness of GAP, it is replaced with a 7×7 convolution, resulting in the model HiFiSeg (w/o GAP). As shown in Table \ref{table3}, the lack of global semantic information leads to a performance decline across all datasets, particularly on the ETIS dataset, where mDice and mIoU drop by 4.1\% and 3.9\%, respectively.

\section{Conclusion}
In this paper, we proposed HiFiSeg network to address the challenges in colon polyp image segmentation, such as fine-grained target localization and boundary feature enhancement. Specifically, the GLIM fused global and local features by extracting multi-scale features in parallel, which facilitated the localization of targets of varying sizes. The SAM selectively combined semantic features with detailed features to alleviate the issue of unclear boundaries, further enhancing performance. Experimental results on five representative colon polyp datasets demonstrated that the HiFiSeg algorithm possessed strong learning and generalization capabilities, outperforming other competing methods. In future work, we plan to explore lightweight architectures to reduce model complexity, thereby extending its applicability to a wider range of medical image segmentation tasks.

\bibliographystyle{IEEEtran}
\bibliography{ref}

\newpage

%If you do not have or do not want to include a photo, you can use IEEEbiographynophoto as shown below:
\begin{IEEEbiographynophoto}
{JINGJING REN} received her bachelor's degree in electronic science and technology from Shanxi Normal University in 2007 and Master's degree in signal and information processing from Chongqing University in 2010. She is currently a lecturer. Her research interests include target detection and image segmentation.
\end{IEEEbiographynophoto}

\begin{IEEEbiographynophoto}{XIAOYONG ZHANG} received his M.S. degree in Signal and Information Processing and his Ph.D. degree in Instrumentation Science and Technology from North Central University in 2010 and 2021, respectively. He is currently an associate professor in the Department of Intelligence and Information Engineering at Taiyuan College. His research interests include embedded signal processing, machine learning, and their applications.
\end{IEEEbiographynophoto}
\begin{IEEEbiographynophoto}{LINA ZHANG.}  received her bachelor's degree in computer science and technology from Northwestern Polytechnical University in 2006 and Master's degree in computer application technology from Taiyuan University of Technology in 2010. She is currently a lecturer. Her research interests include data analysis and target detection.

\end{IEEEbiographynophoto}

\EOD

\end{document}